\documentclass{article}
     \PassOptionsToPackage{numbers, compress}{natbib}

 \usepackage[preprint]{neurips_2021}

\usepackage[utf8]{inputenc} 
\usepackage[T1]{fontenc}    
\usepackage{hyperref}       
\usepackage{url}            
\usepackage{booktabs}       
\usepackage{amsfonts,amssymb}       
\usepackage{nicefrac}       
\usepackage{microtype}      
\usepackage{xcolor}         
\usepackage{marvosym}
\usepackage{graphicx}
\usepackage{diagbox}
\usepackage{multirow}
\usepackage{makecell}

\title{Independent Encoder for Deep Hierarchical Unsupervised Image-to-Image Translation}

\author{%
	Kai Ye, Yinru Ye, Minqiang Yang\textsuperscript{\Letter}, Bin Hu\textsuperscript{\Letter} \\
	Lanzhou University\\
	\texttt{\{kye18, yeyr18, yangmq, bh\}@lzu.edu.cn} \\
}

\begin{document}

\maketitle

\begin{abstract}
The main challenges of image-to-image (I2I) translation are to make the translated image realistic and retain as much information from the source domain as possible. To address this issue, we propose a novel architecture, termed as IEGAN, which removes the encoder of each network and introduces an encoder that is independent of other networks. Compared with previous models, it embodies three advantages of our model: Firstly, it is more directly and comprehensively to grasp image information since the encoder no longer receives loss from generator and discriminator. Secondly, the independent encoder allows each network to focus more on its own goal which makes the translated image more realistic. Thirdly, the reduction in the number of encoders performs more unified image representation. However, when the independent encoder applies two down-sampling blocks, it's hard to extract semantic information. To tackle this problem, we propose deep and shallow information space containing characteristic and semantic information, which can guide the model to translate high-quality images under the task with significant shape or texture change. We compare IEGAN with other previous models, and conduct researches on semantic information consistency and component ablation at the same time. These experiments show the superiority and effectiveness of our architecture. Our code is published on: \url{https://github.com/Elvinky/IEGAN}.
\end{abstract}

\section{Introduction}
Image-to-image (I2I) translation is an essential topic in the field of computer vision. The goal of I2I translation is to learn mutual mapping functions between two different domains with a bijective relationship \cite{zhu2017unpaired,yi2017dualgan}. In recent years, there are mainly two approaches of I2I translation. The first approach is based on supervised learning \cite{mirza2014conditional,Isola_2017_CVPR,Li_2017_NIPS, wang2018high}, it learns mapping functions from paired image sets. However, in many real-world applications, the workload of collecting paired datasets is extremely heavy, so another approach based on unsupervised learning is proposed \cite{Liu_2017_NIPS,Huang_2018_ECCV,Zhu_2017_NIPS}. In this approach, due to the lack of mapping relations of paired samples, it's necessary to use additional rules including weight-coupling\cite{Liu_2017_NIPS, Liu_2016_NIPS}, cycle consistency \cite{zhu2017unpaired,kim2017learning,yi2017dualgan} and identity function \cite{taigman2016unsupervised} to restrict the training of mapping functions.

\begin{figure}
	\includegraphics[width=\textwidth]{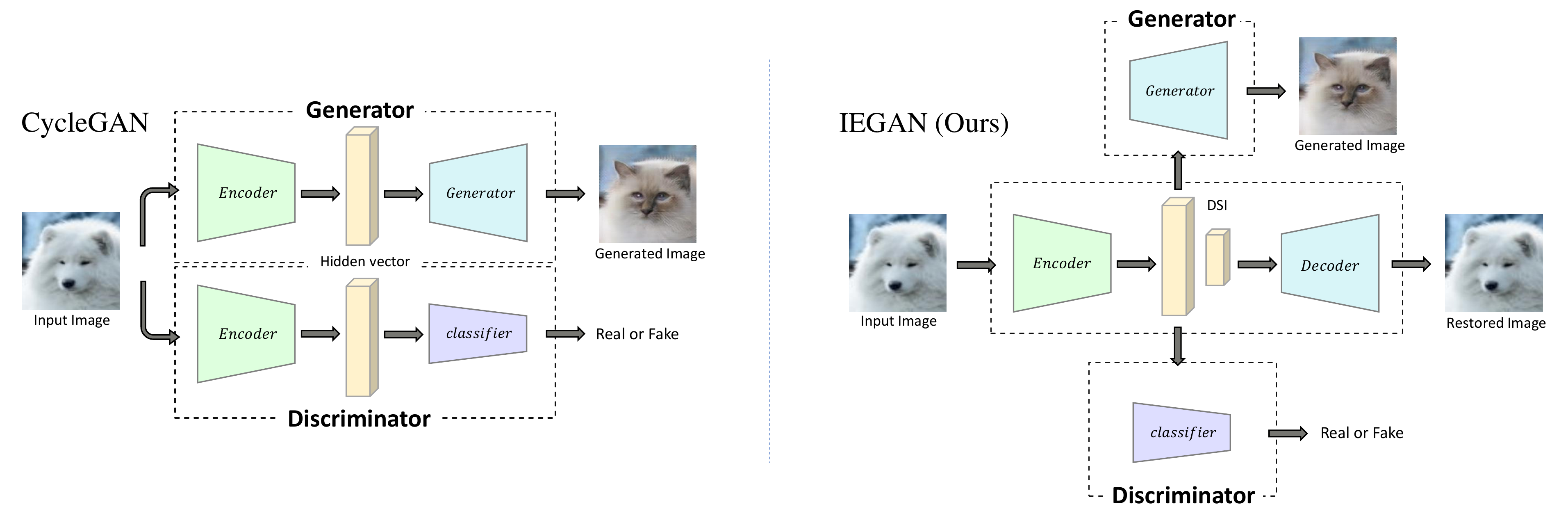}
	\caption{
		The structural differences between previous GAN models, such as CycleGAN(left), and our IEGAN(right) models. DSI means deep and shallow information space.}
	\label{contour_figure}
\end{figure}

Most of I2I translation frameworks are composed of generators and discriminators \cite{zhu2017unpaired,tang2019attention,Kim2020U-GAT-IT:}. Like CycleGAN of Figure \ref{contour_figure}, each generator and dicriminator encodes before translating and classifying respectively. However, the encoder of generator has a bottleneck of encoding capability in I2I translation \cite{Chen_2020_CVPR,radford2015unsupervised}. Compared with the encoder of discriminator directly training through the discrimination loss, gradient received by encoder of generator is back-propagated from the discriminator. This training is indirect for encoder of generator, which causes that the hidden vector learned by an encoder cannot strongly response to the input image. NICE-GAN \cite{Chen_2020_CVPR} proposes a solution by removing the encoder of generator, then generator and discriminator share the same encoder of discriminator.

Reviewing the goal of each component, the goal of an encoder is to learn the hidden vector that can fully represent the features of input image \cite{bengio2013representation}, the decoder can use this hidden vector to restore the input image as much as possible. The ultimate goal of discriminator is to distinguish between the translated image from the source domain and the real image from the target domain \cite{Ian_2014_NIPS}. However, discriminators and encoders have different goals, which makes the hidden vector learned by the encoder of discriminator well complete the task of classification but not suitable for task of generation. This is because information that is not conducive to classification won't be learned into the hidden vector.

In view of above mentioned reasons, we propose a novel architecture which refines the goal of each network, as illustrated in Figure \ref{contour_figure}. Specifically, we remove the encoders of generators and discriminators, and introduce an independent encoder, which means that the encoder is no longer affected by other networks. In other words, the generator and discriminator don't need to encode before achieving the goal, and the training of encoder is also independent of the training of other networks. Such kind of independence exhibits three advantages: \textbf{I.} Independent encoder can grasp image information more directly and comprehensively. Because this architecture guides the encoder to focus on learning input image representation and ignore the goals of other networks.  \textbf{II.} The translated image is of higher quality and retains more information from the source domain. \textbf{III.} With reduction in the number of encoders in Figure \ref{contour_figure} decreasing, the number of image representations required for the model is also reduced, which brings the more unified representation.

The performance of previous methods depends on the amount of changes of shape and texture between domains. When independent encoder applies two down-sampling blocks \cite{kim2017learning,lee2018diverse}, the style transfer can be successfully performed. But it's strenuous to complete tasks with significant shape changes (e.g. The cat is translated into the dog). The hidden vector learned by independent encoder only contains characteristic information (e.g. color and texture) of the input image \cite{Wang_2020_NIPS}. To mitigate this problem, we propose deep and shallow information space (DSI) composed of different layers of hidden vectors. In an unsupervised environment, the model obtains the DSI of the input image through an independent encoder, and then asks the decoder to use DSI to restore exactly the same input image. At the same time, DSI merges and superimposes the hidden vectors of different layers and transmits them to the generator and discriminator. In this way, generator and discriminator can use characteristic and semantic information to complete task with significant shape or texture change.

We perform experiments on several popular benchmarks on multiple datasets. Our method outperforms various state-of-the-art counterparts. We further evaluate the independent encoder through semantic information consistency which proves the ability of each model to retain source domain information. In the meantime, ablation studies are conducted to verify the effectiveness of each component.

\section{Related work}
\label{related_work}

\paragraph{Generative adversarial networks.} GAN \cite{Ian_2014_NIPS} has done a large number of practical use cases, such as image generation \cite{Zhang_2017_ICCV}, artwork generation \cite{tan2017artgan}, music generation \cite{engel2019gansynth}, and video generation \cite{Tulyakov_2018_CVPR}. In addition, it can also improve image quality \cite{tong2017image}, image coloring \cite{Zhang_CORR_2016}, face generation \cite{Karras_2019_CVPR}, video encoding \cite{wang2020one}, and other more interesting tasks. GAN has several approaches to improve the authenticity of translated images. The first approach is to improve training stability (e.g. DCGAN \cite{radford2015unsupervised} used stride convolution and transposed convolution to improve training stability). The second one is large-scale training (e.g. BIGGAN \cite{Borck_2018_CORR} synthesized realistic images by increasing batchsize and truncation techniques). The third one is architectural modifications(e.g. SAGAN \cite{zhang2019self} added self-attention mechanism to the network). The GAN models mentioned above are all based on probability models, but there are some GAN models based on energy models, such as EBGAN \cite{zhao2016energy} and BEGAN \cite{berthelot2017began}. These models are classified by the reconstruction of input image through the discriminator composed of encoder-decoder.

\paragraph{Unsupervised I2I translation.} I2I translation based on unpaired datasets has been widely studied in the field of computer vision since CycleGAN \cite{zhu2017unpaired} was proposed. U-GAT-IT \cite{Kim2020U-GAT-IT:} added AdaILN and CAM \cite{zhou2016learning} to the generator and the discriminator. DeepI2I \cite{Wang_2020_NIPS} used a pre-trained discriminator as the encoder of the generator. And NICE-GAN \cite{Chen_2020_CVPR} removed the encoder of the generator and used the encoder of the discriminator. These are optimizations for the generator and the discriminator. MUNIT \cite{Huang_2018_ECCV} and DRIT \cite{lee2018diverse} decoupled the latent space of images into information of content and style. UNIT \cite{Liu_2017_NIPS} and ComboGAN \cite{anoosheh2018combogan} used domain to share latent space. These are researches on image representation. UPD \cite{yi2020unpaired} and GANILLA \cite{hicsonmez2020ganilla} focused on image translation under specific tasks in generating artistic portrait line drawings and illustrations. StarGAN \cite{Choi_2018_CVPR} achived multi-domains image translation by adding mask vector to domain labels.

\paragraph{Representation learning.} The researches and developments of unsupervised representation learning includes probabilistic models \cite{salakhutdinov2009deep}, autoencoders \cite{hinton2006reducing}, and deep networks. The goal of representation learning is to find a better way to express data \cite{bengio2013representation}. A good data representation will greatly improve the efficiency of the model. Representation learning can be implemented in three ways: supervised learning, self-supervised learning and unsupervised learning. Unsupervised learning uses an encoder-decoder network to perform dimensionality reduction and compression on the input data, thereby discarding redundant information, and selecting the most critical concentrated information. In the deep network, the encoder-decoder network constitutes a powerful method of representation learning. 

\section{Our approach}
\begin{figure}
\includegraphics[width=\textwidth]{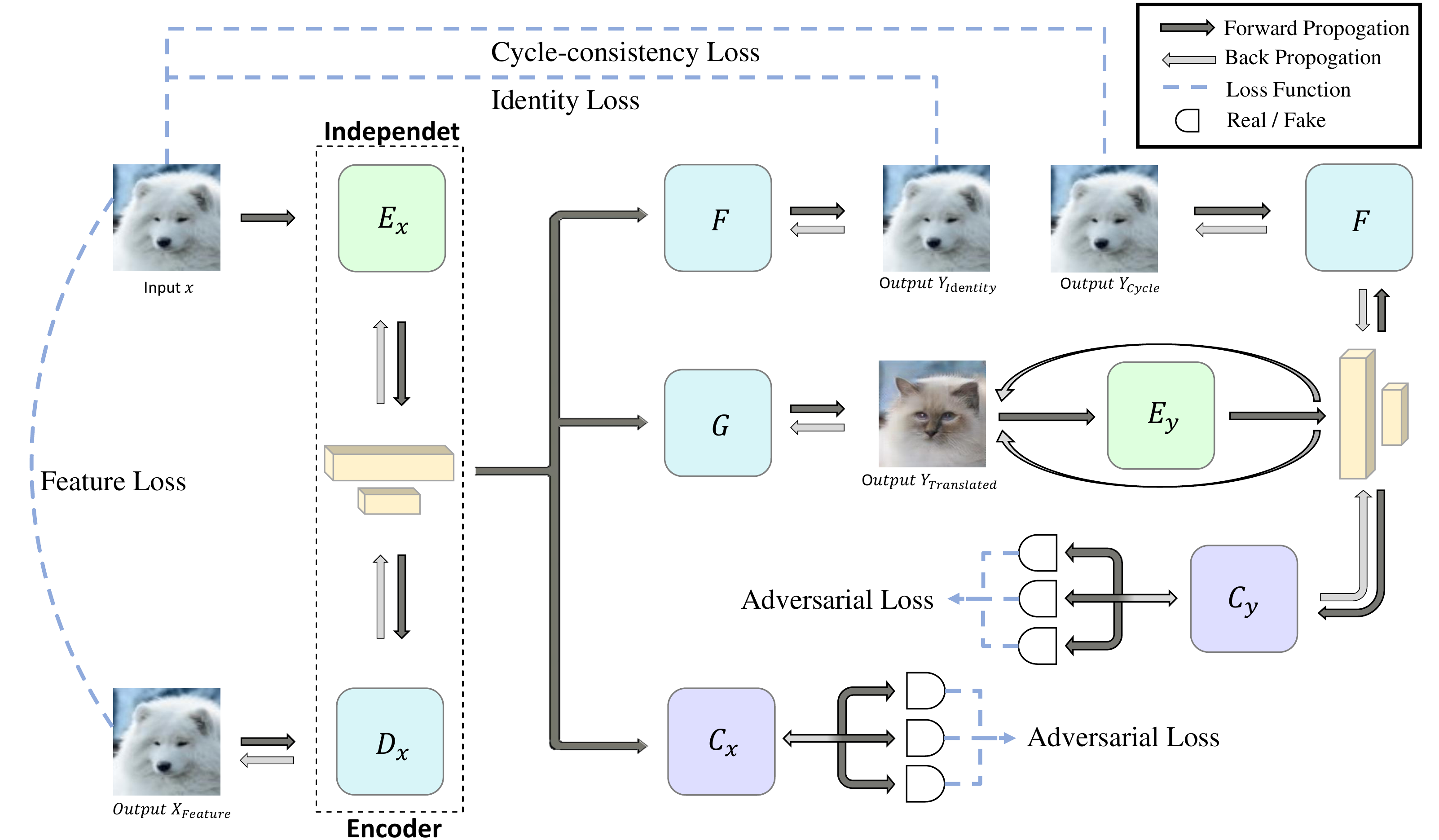}
\caption{
	\textbf{The structure flowchart of IEGAN.} The structural and purposeful differences between previous GAN models such as CycleGAN(left) and our IEGAN(right) models, DSI means deep and shallow information.}
\label{struct}
\end{figure}

\subsection{Overview}
\paragraph{Problem definition.} $x\in\{\mathcal{X}\}$ and $y\in\{\mathcal{Y}\}$ are samples respectively in the source domain and target domain. In I2I translation, the ultimate goal is to train the mapping function of $G:x\to y$ and the inverse mapping function of $F:y\to x$ \cite{zhu2017unpaired}. In order to train them, some additional constraints are neccessary. For example, in supervised learning, after a paired dataset $\{x_{i},y_{i}\}_{i=1}^{N}$ given, mapping functions are restricted by conditions of $G(x_{i})=y_{i}$ and $F(y_{i})=x_{i}$ \cite{Isola_2017_CVPR}. 

But in unsupervised learning, there are only unpaired datasets $\{x_{i}\}_{i=1}^{N}$ and $\{y_{j}\}_{j=1}^{M}$. Without  restriction of the pairing relation, the function can be mapped to any distributions. To tackle this issue, previous methods put forward additional conditions, such as cycle-consistency $F(G(x_{i}))=x_{i}$ (resp.$G(F(y_{j}))=y_{j}$) \cite{zhu2017unpaired,kim2017learning,yi2017dualgan} and identity-mapping-enforcing $G(y_{j})=y_{j}$ (resp.$F(x_{i})=x_{i}$) \cite{taigman2016unsupervised,zhu2017unpaired}. 

Even if the above conditions guide model to learn mapping functions between two domains, the translated images based on the mapping functions trained by the above conditions are blurred. In other words, the translated image is a mixture of multiple distributions \cite{Ian_2014_NIPS}. So GAN introduces two discriminators $C_{x}$ and $C_{y}$, where $C_{x}$ (resp. $C_{y}$) calculates the conditional probability that the sample $F(y_{j})$ (resp. $G(x_{i})$) matching to the distribution $\mathcal{X}$ (resp. $\mathcal{Y}$).

\paragraph{Independent encoder in I2I translation.} In the field of I2I translation, the common GAN models consists of four core networks \cite{Chen_2020_CVPR,Kim2020U-GAT-IT:,yi2020unpaired}: two generators $G$ (consists of $E_{x}^{G}$ and $G$) and $F$ (consists of $E_{y}^{F}$ and $F$), two discriminators $C_{x}$(consists of $E_{x}^{D}$ and $C_{x}$) and $C_{y}$ (consists of $E_{y}^{D}$ and $C_{y}$). And an encoder is embedded in each core network. We have observed two phenomenas: \textbf{I.} Because the input of each network is an image and there are only two image domains, we don’t need so many encoders, only two encoders $E_{x}$ and $E_{y}$ \textbf{II.} The generator needs to encode the image first and then translate it. Also the discriminator encodes first and then classifies the image. However, the generator and discriminator can actually be used without encoding.

As mentioned above, if the encoder is embedded in other networks, the encoding capability of encoder will be limited by the network which it belongs to. This is because training of encoder depends on the back-propagation gradient received by the network. And the gradient depends on the loss function known as the goal of the network. \cite{Ian_2014_NIPS} The purpose of encoder is to learn the DSI of the input image, and the goal of discriminator is to map the input image into vector to determine whether the domains are aligned \cite{Chen_2020_CVPR}. The vector refers to the output of the generator or the target domain, and the goal of the generator is to translate an image in another domain \cite{zhu2017unpaired}, which means the goal of the encoder is different from the goals of generator and discriminator. If the encoder is in the discriminator, the target of  encoder becomes classification, which causes the DSI to lose information irrelevant to classification. If the encoder is in the generator, the loss of encoder is generated by another network. This is an indirect way which causes the DSI not respond strongly to the input image.

So we propose to remove the encoder $E_{x}$ of each network and establish an encoder-decoder ($E_{x}$ and $D_{x}$) as an independent encoder to be trained, Figure \ref{struct} illustrates our architecture. The training of the encoder now no longer receives the loss generated by other networks but instead receives the feature loss (Eq. \ref{feature_loss}) between the output image of the decoder $D_{x}(E_{x}(x))$ and the input image $x$. In this way, the encoder ignores the goal of the generator or discriminator, and focuses on learning the DSI of the input image, thereby ensuring the encoding capacility of encoder. At the same time, the encoder transmits DSI to the generator $G$ and discriminator $C_{x}$. When the encoder applies feature loss, DSI also contains more information about the input image. Such an encoder can improve the generating ability of the model, which makes it easier to translate high-quality images.

\begin{figure}
\centering
\includegraphics[width=\textwidth]{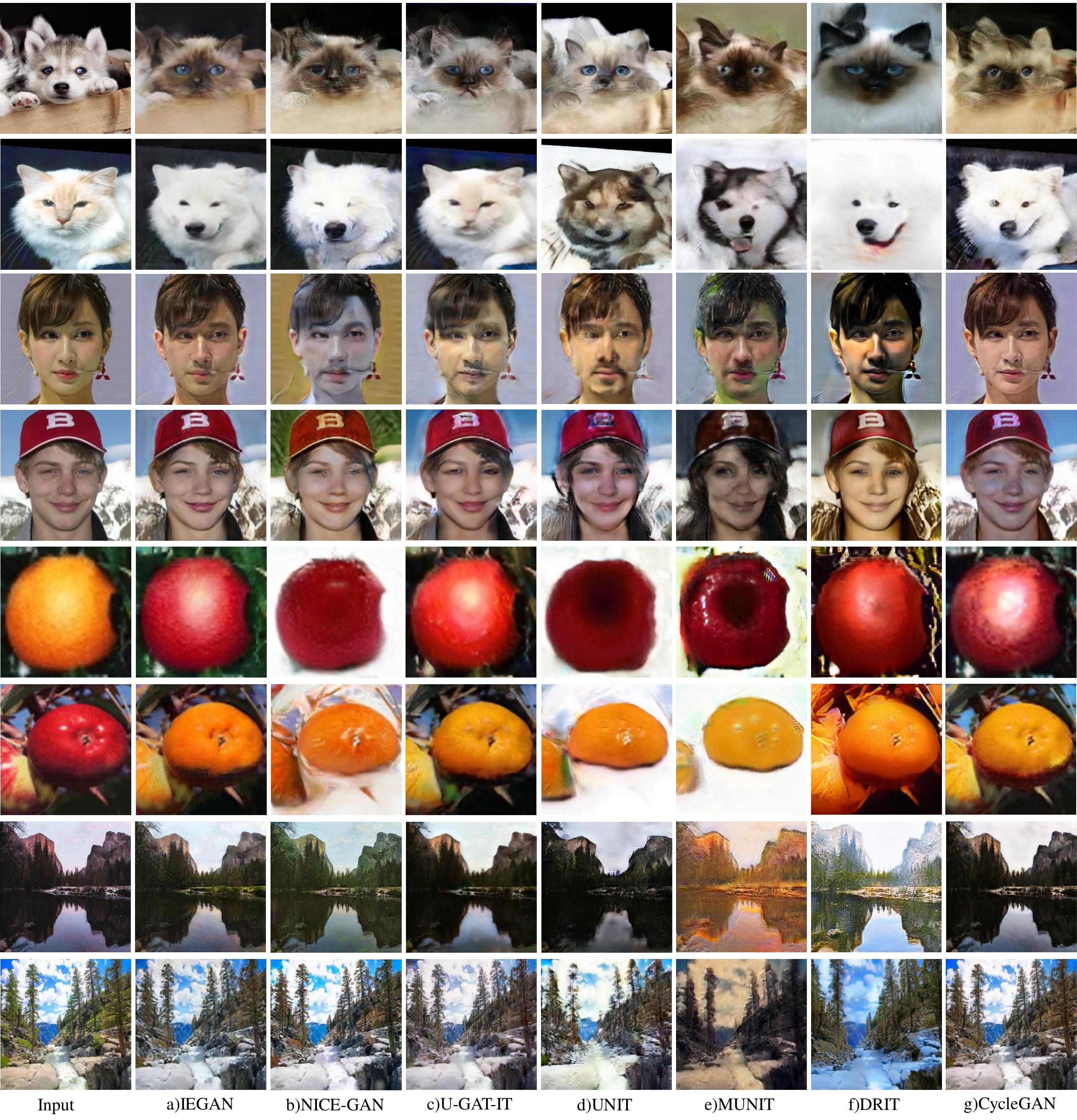}
\caption{
	\textbf{Examples of translated output on four datasets.} From top to bottom: \emph{cat2dog},  \emph{man2woman},  \emph{apple2orange} and  \emph{summer2winter}.}
\label{compare}
\end{figure}

\subsection{Architecture}

\paragraph{Independent encoders $E_{x}$ and $E_{y}$.}
The idea of an independent encoder was inspired by previous work \cite{berthelot2017began,ronneberger2015u,hinton2006fast,HE_2015_CORR}. In these previous works, encoder-decoder network is a powerful method in unsupervised representation learning, the encoder progressively downsamples the input to obtain image information. The decoder uses the learned image information to progressively upsample and restore the input image. When the number of down-sampling blocks is small, the encoder can only obtain characteristic information. When the number of down-sampling blocks increases, even if encoder obtains semantic information, it is difficult for the decoder to restore a image similar to the input image \cite{Wang_2020_NIPS}. 

By adding the skip connections between encoder and decoder to construct an U-net \cite{ronneberger2015u} network, the above problems can be solved ingeniously. In this paper, we apply U-net to build an independent encoder, which means that the independent encoder includes an encoder and a decoder. In order to improve the encoding capability of the encoder, we add linear attention transformer \cite{kitaev2020reformer} in the encoder.

Most of I2I models only transmit the characteristic information of the input image to generator and discriminator. There are also networks that do other processing, such as decoupling the latent space into content information and style information \cite{Huang_2018_ECCV}. As the resolution increases,  image information is transmited to the generator through the adapter network layer by layer \cite{Wang_2020_NIPS}. We propose an approach that is different from previous models. We extract the information of the input image to DSI, which contains $\mathbb{R}^{H\times W\times C}=\mathbb{R}^{(ch*2^{n})\times (ch*2^{n})\times (ch*2^{-n})}, n\in \{-3,-2,-1,0\},ch=64.$ Then the DSI merges and superimposes the hidden vectors of different layers into $\mathbb{R}^{H\times W\times C}=\mathbb{R}^{(ch*2^{m})\times (ch*2^{m})\times (ch*2^{-2m})},m=1,$ during the up-sampling process, model transmits it to other networks.

\paragraph{Generator and discriminator.}
Generators $G$ and $F$ are composed of U-GAT-IT-light \cite{Kim2020U-GAT-IT:} generators without encoders. The generator contains six layers of residual blocks with AdaILN and two sub-pixel up-sampling layers. The discriminators $C_{x}$ and $C_{y}$ are composed of an attention mechanism CAM \cite{zhou2016learning} and a down-sampling network with a multi-scale mechanism \cite{durugkar2016generative}.
\begin{table}
\centering
\caption{\textbf{Comparison with baselines.} FID and KID$\times$100 for different algorithms. Lower is better.}
\resizebox{\textwidth}{0.32\textwidth}{
	\begin{tabular}{|c|c|c|c|c|c|c|c|c|} 
		\hline
		\multirow{2}{*}{\diagbox{Method}{Dataset}} & \multicolumn{2}{c|}{dog$\to$cat}   & \multicolumn{2}{c|}{woman$\to$man}   & \multicolumn{2}{c|}{orange$\to$apple} & \multicolumn{2}{c|}{winter$\to$summer}  \\ 
		\cline{2-9}
		& FID            & KID$\times$100     & FID            & KID$\times$100     & FID             & KID$\times$100       & FID             & KID$\times$100        \\ 
		\hline\hline
		\makecell*[l]{\textbf{IEGAN}}                                      & \textbf{43.54} & \textbf{0.95} & \textbf{138.29} & \textbf{3.56} & \textbf{132.86} & \textbf{4.65}   & 83.35 & 1.76    \\ 
		\hline
		\makecell*[l]{NICE-GAN}                                   & 48.79          & 1.58          & 145.31         & 4.28          & 169.79          & 8.00            & \textbf{76.44}  & \textbf{1.22}             \\ 
		\hline
		\makecell*[l]{U-GAT-IT-light}                             & 80.75          & 3.22          & 159.73         & 5.75          & 134.20          & 4.92            & 80.33           & 1.82             \\ 
		\hline
		\makecell*[l]{CycleGAN}                                   & 119.32         & 4.93          & 146.97         & 4.84          & 142.47          & 5.46            & 79.58           & 1.36             \\ 
		\hline
		\makecell*[l]{UNIT}                                       & 59.56          & 1.94          & 179.56         & 9.18          & 161.25          & 7.60            & 95.93           & 4.63             \\ 
		\hline
		\makecell*[l]{MUNIT}                                      & 53.25          & 1.26          & 163.39         & 6.68          & 186.88          & 9.09            & 99.14           & 4.66             \\ 
		\hline
		\makecell*[l]{DRIT}                                       & 94.50          & 5.20          & 187.15         & 10.00          & 141.07          & 6.17            & 78.61          & 1.69             \\
		\hline
		\hline
		\multirow{2}{*}{\diagbox{Method}{Dataset}} & \multicolumn{2}{c|}{cat$\to$dog}   & \multicolumn{2}{c|}{man$\to$woman}   & \multicolumn{2}{c|}{apple$\to$orange} & \multicolumn{2}{c|}{summer$\to$winter}  \\ 
		\cline{2-9}
		& FID            & KID$\times$100     & FID            & KID$\times$100     & FID             & KID$\times$100       & FID             & KID$\times$100        \\ 
		\hline\hline
		\makecell*[l]{\textbf{IEGAN}}                                      & \textbf{38.68} & \textbf{0.67} & \textbf{136.44} & 3.91 & \textbf{164.58} & \textbf{8.70}   & 97.71 & 2.07    \\ 
		\hline
		\makecell*[l]{NICE-GAN}                                   & 44.67          & 1.20          & 139.30         & \textbf{3.12} & 192.19          & 11.67            & \textbf{76.03} & \textbf{0.67}            \\ 
		\hline
		\makecell*[l]{U-GAT-IT-light}                             & 64.36          & 2.49          & 151.06         & 4.31         & 179.29          & 10.17            & 88.41           & 1.43            \\ 
		\hline
		\makecell*[l]{CycleGAN}                                   & 125.30         & 6.93          & 161.62         & 5.64         & 181.60          & 10.42            & 78.76           & 0.78                 \\ 
		\hline
		\makecell*[l]{UNIT}                                       & 63.78          & 1.94          & 147.81         & 4.81         & 195.24          & 12.33            & 112.07          & 5.36            \\ 
		\hline
		\makecell*[l]{MUNIT}                                      & 60.84          & 2.42          & 150.94         & 5.62          & 213.77          & 14.04            & 114.08          & 5.27            \\ 
		\hline
		\makecell*[l]{DRIT}                                       & 79.57          & 4.57          & 141.13         & 3.53          & 176.13          & 10.10            & 81.64          & 1.27             \\
		\hline
\end{tabular}}

\label{fid_kid}
\end{table}

\subsection{Loss functions}
The training process is consisted of four types of losses: feature loss, adversarial loss, identity reconstruction loss, and cycle-consistency loss. We explain them in detail as follows:

\emph{\textbf{Feature loss.}}\quad We use average absolute loss to ensure that there is a stable gradient in any input situation to provide a higher quality and more accurate encoding capability. \textbf{Except for this loss, the encoders remain unchanged in the following three losses.} \begin{equation} \min_{E_{x}}L_{fea}^{x}=\mathbb{E}_{x\sim\mathcal{X}}[|x-D_{x}(E_{x}(x))|_{1}]. 
\label{feature_loss}
\end{equation}

\emph{\textbf{Adversarial loss.}}\quad The adversarial loss guides the discriminator to distinguish image between source domain and target domain, and make the distribution probability of the output of the generator continuously approach of target domain. \begin{equation} \min_{G}\max_{C_{y}}L_{adv}^{x\to y}=\mathbb{E}_{y\sim\mathcal{Y}}[(C_{y}(E_{y}(y)))^{2}]+\mathbb{E}_{x\sim\mathcal{X}}[(1-C_{y}(G(E_{x}(x))))^{2}]. 
\end{equation}

\emph{\textbf{Cycle-consistency loss.}}\quad In order to prevent mode collapse in I2I translation, we apply cycle-consistency loss to have input image translated into target domain, and then translated image back to source domain, which the translated image should be consistent with the input image. \begin{equation} \min_{G,F}L_{cyc}^{x\to y}=\mathbb{E}_{x\sim\mathcal{X}}[|x - F(E_{y}(G(E_{x}(x))))|_{1}]. 
\end{equation}

\emph{\textbf{Identity loss.}}\quad To address the steganography issue \cite{chu2017cyclegan}, we apply identity loss to ensure that $G$ and $F$ no longer have a similar behavior, which is decryption and encryption for hidden vector, so the output distribution of generator can be closer to the distribution of target domain. \begin{equation} \min_{F}L_{ide}^{x\to y}=\mathbb{E}_{x\sim\mathcal{X}}[|x-F(E_{x}(x))]|_{1}]. 
\end{equation}

\emph{\textbf{Full objective.}}\quad In summary, final objective of our model is optimized by jointly training independent encoders, generators, and discriminators. \begin{equation} \min_{G,F,E_{x},E{y}}\max_{C_{x},C_{y}}\lambda_{1}L_{fea}+\lambda_{2}L_{adv}+\lambda_{3}L_{cyc}+\lambda_{4}L_{ide},
\end{equation} 
where, $\lambda_{1}=10$, 
$\lambda_{2}=1$, $\lambda_{3}=10$, $\lambda_{4}=10$. Here, $L_{fea}=L_{fea}^{x}+L_{fea}^{y}$, $L_{adv}=L_{adv}^{x\to y}+L_{adv}^{y\to x}$, $L_{cyc}=L_{cyc}^{x\to y}+L_{cyc}^{y\to x}$ and $L_{ide}=L_{ide}^{x\to y}+L_{ide}^{y\to x}$.
\begin{figure}
\centering
\includegraphics[width=\textwidth]{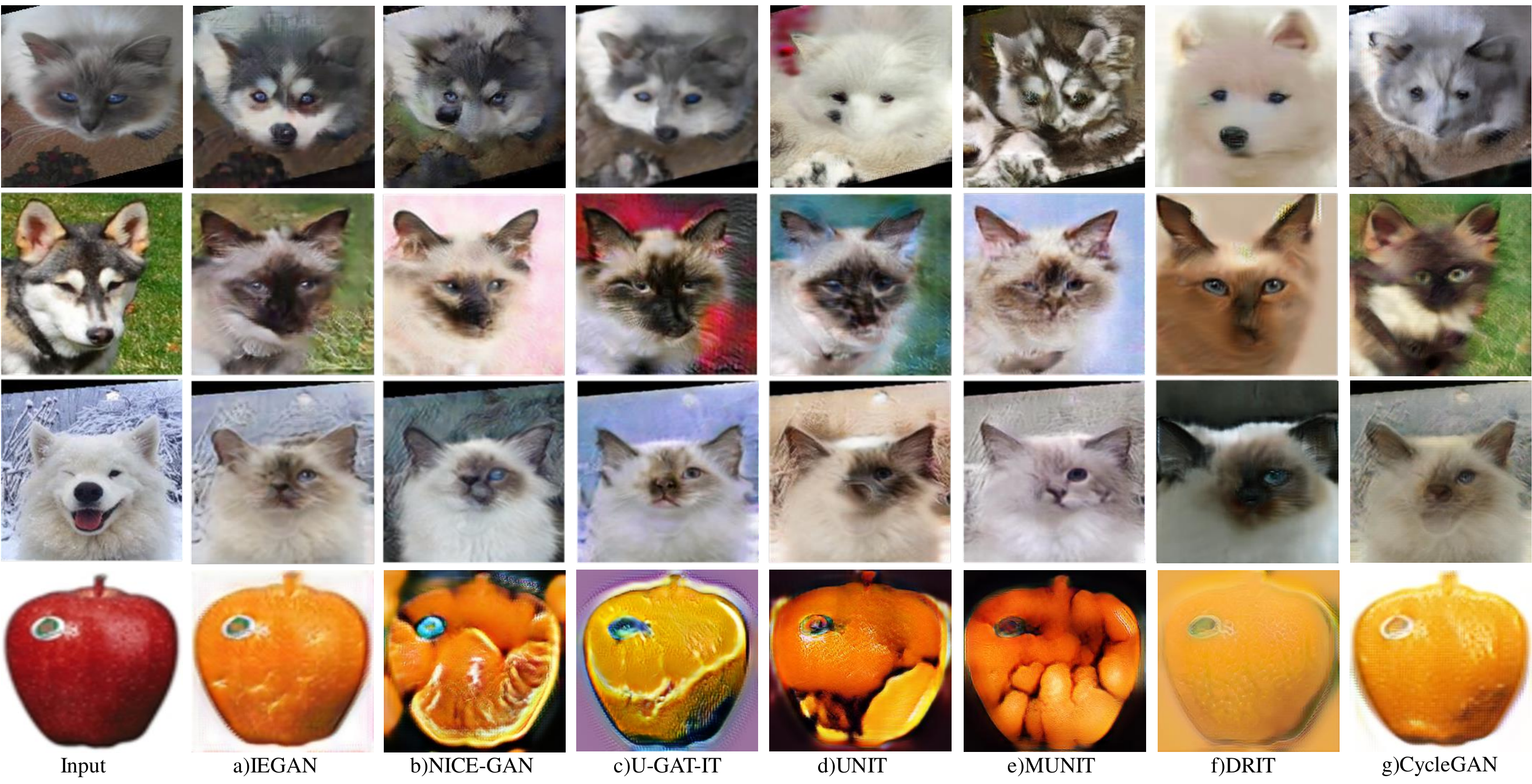}
\caption{
	\textbf{Semantic information consistency.} Compared with other models, IEGAN can better retain the background information of the image under the premise of translating high-quality images.}
\label{semantic_consistency}
\end{figure}
\section{Experiments}

\subsection{Experimental setup} \label{setup}
All metrics are obtained through 100K iterations of training on the NVIDIA Tesla P100 GPU. We use the Adam optimizer with $1\times10^{-4}$ weight decay and learning rate, and apply ReLU and leaky-ReLU with a slope of $0.2$ as the activation functions of the generator and discriminator respectively. We resize the image to $286\times286$ and randomly crop to $256\times256$ for data  augmentation.
\paragraph{Baselines.} We compare the performance of metrics of our method with six different state-of-the-art I2I translation methods including NICE-GAN \cite{Chen_2020_CVPR}, U-GAT-IT \cite{Kim2020U-GAT-IT:},  UNIT \cite{Liu_2017_NIPS}, MUNIT \cite{Huang_2018_ECCV}, DRIT \cite{lee2018diverse} and CycleGAN \cite{zhu2017unpaired}, all of which achieves translation between different domains. All models are implemented by the public code on github.

\paragraph{Datasets.} We consider four unpaired datasets: \emph{cat2dog}, \emph{man2woman}, \emph{apple2orange}, \emph{summer2winter}. We crop and resize all input images to 256$\times$256 during training, and the output images are also 256$\times$256. The number of splits for the training set and test set of all datasets is based on the following template(train$\mathcal{X}$-$\mathcal{Y}$/test$\mathcal{X}$-$\mathcal{Y}$): \emph{cat2dog} (771-1264/100-100); \emph{man2woman} (1200-1200/115-115); \emph{apple2orange} (995-1019/266-248); \emph{summer2winter} (1231-962/309-238). \emph{cat2dog} dataset is studied in DRIT, \emph{apple2orange} and \emph{summer2winter} dataset are studied in CycleGAN. We created \emph{man2woman} dataset to randomly filter images after classification by gender on FFHQ \cite{Karras_2019_CVPR}.

\paragraph{Evaluation metrics.}
For quantitative evaluation, we choose Fréchet Inception Distance \cite{Heusel_2017_NIPS} (FID) and Kernel Inception Distance \cite{Binkowski_2018_ICLR} (KID)  as evaluation metrics. FID calculates the distance between real and translated images in hidden vector given by the features of a convolutional neural network. KID computes the squared Maximum Mean Discrepancy to get the visual similarity of real and translated images. KID has an unbiased estimator, which is more consistent with human evaluation.
Low values of FID and KID scores mean the excellent perfomance of GAN. 
\subsection{Results}
\paragraph{Comparisons with state of the arts.}
We first test IEGAN on four datasets, using six baselines for comparison. We report our key results in Table\ref{fid_kid} and Figure \ref{compare}.

\emph{Visual analysis.}\quad
From the perspective of visual analysis, Figure \ref{compare} shows examples of image translation of different models on different datasets. In general, the images translated by IEGAN are more difficult to distinguish from the images in the target domain, which shows that our model has excellent translation capability at a qualitative level. Both shape and texture are important bases for human perception of images. In terms of shape, the shapes of animals, people, fruits, and landscapes translated by IEGAN are closer to realistic images. For example, facial features of the translated cat in the first row of Figure \ref{compare} are more realistic than other models. In terms of texture, images translated from models under different architectures will have artifacts to a certain extent. In contrast, our model can better reduce the appearances of artifact.

\emph{Metric anaysis.}\quad \label{limit}
From metric analysis, Table \ref{fid_kid} shows the FID and KID scores of the above models on the four datasets. In brief, except for \emph{summer2winter} dataset, the metrics of IEGAN  achieve the lowest scores on all datasets and have a significant reduction of FID and KID, especially KID. This shows that our model has an excellent translation capability at a quantitative level. For example, on the popular \emph{cat2dog} dataset, the best KID score of $dog\to cat$ we obtained was $0.95$, which is $0.63$ lower than NICE-GAN, and the FID score also dropped from $48.79$ to $43.54$. Compared to some models that only obtain good scores on the mapping function in one direction, we also reduced the KID from $1.20$ to $0.67$ on $cat\to dog$, and the FID from $44.67$ to $38.68$. We can also notice from the Table \ref{fid_kid} that CycleGAN can cope with the task of texture change well. UNIT and MUNIT can achieve the goal of shape change. The images translated by DRIT is real but sometimes have nothing to do with input images. The images translated by NICE-GAN and U-GAT-IT have fewer artifacts.
\begin{table}
\centering
\caption{\textbf{Ablation study.} IE: independent encoder; DSI: deep and shallow information space; LAT: add linear attention transformer in IE; $\times$ in Components: removing this component; $\checkmark$ in Components: containing this component.}
\begin{tabular}{|c|c|c|c|c|c|c|} 
	\hline
	\multicolumn{3}{|c|}{Components} & \multicolumn{2}{c|}{dog$\to$cat} & \multicolumn{2}{c|}{cat$\to$dog}  \\ 
	\hline\hline
	\makecell*[l]{IE} & DSI & LAT                   & FID   & KID $\times$ 100            & FID   & KID $\times$ 100             \\
	\hline
	\makecell*[l]{$\times$}  & $\times$   & $\times$                     & 54.98     & 1.85                   &   51.93    & 1.38                      \\
	\makecell*[l]{$\checkmark$}  & $\times$   & $\times$                     & 47.00 & 1.36                 & 40.81 & 0.87                  \\
	\makecell*[l]{$\checkmark$}  & $\checkmark$   & $\times$                     & 45.62 & \textbf{0.92}                & 39.74 & 0.79                  \\
	\makecell*[l]{$\checkmark$}  & $\times$   & $\checkmark$                     & 47.80     & 0.93                    &  40.85     &     1.06                  \\
	\makecell*[l]{$\checkmark$}  & $\checkmark$   & $\checkmark$                     & \textbf{43.54} & 0.95                 & \textbf{38.68} & \textbf{0.67}                  \\
	\hline
\end{tabular}

\label{ablation}
\end{table}
\paragraph{Semantic information consistency.}
In the process of I2I translation, not every part of image needs to be translated. We refer to the parts that need to be translated and don't need to be translated as subject and background respectively. Even though most of time we only translate the subject, the background is still a part of the image information. The loss of background information from the source domain to the target domain is also a kind of semantic information inconsistency. The semantic information consistency is reflected in the shape, texture, and color of background information. For instance, in the first row of Figure \ref{semantic_consistency}, a cat is on a table with patterns. After translation, it should be a dog on a table with patterns. In the same example, in the fourth row of Figure \ref{semantic_consistency}, there is a green label on the apple. After translation, there should be a green label on the orange. Compared with other models, it is obvious that our model has semantic information consistency.

Under tasks of I2I translation, FID and KID are high-quality metrics for evaluating the quality and diversity of translated images, but they cannot fully explain the completeness of translation work. In other words, it is important to translate realistic images, but if semantic information consistency is lost in the translation process, the translation relationship between output and input is weakened \cite{Kim2020U-GAT-IT:}. For IEGAN, because the goal of an independent encoder is to focus on learning the DSI of input image, it can preserve as much background information of input image as possible to the generator on the premise of translating high-quality images in order to achieve semantic information consistency.
\paragraph{Ablation study.}
In the Table \ref{ablation}, we next compare the individual impact of each proposed component of the independent encoder in GAN model on the \emph{cat2dog} dataset, and compare the FID and KID scores. We analyze three key components including IE, DSI and LAT. Table \ref{ablation} shows that the application of IE reduces FID and KID by $15\%$ to $37\%$, which means that an independent encoder can obtain better image representation while also allowing each network to focus on its own goal. This strategy of using independent encoder can improve translation capability of model. In further analysis of DSI, compared with IE, the application of DSI doesn't significantly optimize the quality of the translated image but still lower every metric, which proves the effectiveness of DSI and also means that DSI can make better use of hierarchical information to provide more information to other networks, which makes the quality of translated images better. Finally, we notice that using LAT alone won't produce positive effects, but combined with DSI, it can significantly improve the translation quality of the model. Overall, by merging all components, IEGAN is significantly better than all other variants.

\section{Conclusion}
In this paper, we propose a novel unsupervised I2I translation architecture, called IEGAN. This architecture allows the encoder to be independent of generator and discriminator. It can more directly and comprehensively grasp the image information. In addition, we introduce a deep and shallow information space based on deep hierarchical I2I translation. The proposed representation allows other networks to obtain the characteristic and semantic information of the input image, which makes the translated image more realistic. Through experiments, we have proved that our model is more superior and effective than previous models.

\section*{Broader impact}\label{impact}
Our model is mainly used to translate images. The translated images can be applied to business models such as movies, advertisements, games and even virtual reality. Also, this technology can be used to solve the automatic translation of faces or objects. Traditionally, it's a labor-intensive technology, which means that the emergence of this technology has lowered the application threshold for \emph{deepfakes}. Non-professionals can also use this technology to fake information, which may harm the rights of individuals. Seriously, it may endanger the safety of enterprises and the country.

\bibliographystyle{plain}
\bibliography{ref} 

\section*{Checklist}
\begin{enumerate}
	
	\item For all authors...
	\begin{enumerate}
		\item Do the main claims made in the abstract and introduction accurately reflect the paper's contributions and scope?
		\answerYes{} \emph{The scope of this paper is unsupervised I2I translation, and the main contribution is a novel translation architecture, which are shown in the abstract and introduction.}
		\item Did you describe the limitations of your work?
		\answerYes{} \emph{We pointed out in section\ref{limit} of the paper that our model doesn't perform well in the clarity of the translated images on the summer2winter dataset.}
		\item Did you discuss any potential negative societal impacts of your work?
		\answerYes{} \emph{We discussed the impact of our work on the future, see section\ref{impact} for details.}
		\item Have you read the ethics review guidelines and ensured that your paper conforms to them?
		\answerYes{} \emph{We carefully read the ethical review guidelines and ensured that our papers meets these guidelines.}
	\end{enumerate}
	
	\item If you are including theoretical results...
	\begin{enumerate}
		\item Did you state the full set of assumptions of all theoretical results?
		\answerYes{} \emph{The assumption of the work of image translation is that there needs to be a bijective relationship between domains, which can be seen in the introduction. For example, the translation from cat to chair and human face to orange is meaningless.}
		\item Did you include complete proofs of all theoretical results?
		\answerYes{} \emph{We compared other models and conducted other experiments to prove our theoretical results.}
	\end{enumerate}
	
	\item If you ran experiments...
	\begin{enumerate}
		\item Did you include the code, data, and instructions needed to reproduce the main experimental results (either in the supplemental material or as a URL)?
		\answerYes{} \emph{The code, data, and instructions to reproduce our experiment are published on the github via URL in abstract.}
		\item Did you specify all the training details (e.g., data splits, hyperparameters, how they were chosen)?\answerYes{} \emph{We give all the training details of the experiment in section\ref{setup}.}
		\item Did you report error bars (e.g., with respect to the random seed after running experiments multiple times)?
		\answerNo{} \emph{The error of the experiment was mainly caused by the order after the training set was shuffled, but after 100K iterations training, the error will not deviate much. So we did not give the error bars.}
		\item Did you include the total amount of compute and the type of resources used (e.g., type of GPUs, internal cluster, or cloud provider)?
		\answerYes{} \emph{In section\ref{setup}, we give the experimental environment and resource consumption.}
	\end{enumerate}
	
	\item If you are using existing assets (e.g., code, data, models) or curating/releasing new assets...
	\begin{enumerate}
		\item If your work uses existing assets, did you cite the creators?
		\answerYes{} \emph{We have cited the creators for all data and codes used.}
		\item Did you mention the license of the assets?
		\answerNo{} \emph{We checked the licenses and cited the datasets. Due to space limitation, those kinds of studies do not state the license information.}
		\item Did you include any new assets either in the supplemental material or as a URL?
		\answerYes{} \emph{We publish the code on the github via URL in abstract.}
		\item Did you discuss whether and how consent was obtained from people whose data you're using/curating?
		\answerYes{} \emph{All the data we use are open dataset.}
		\item Did you discuss whether the data you are using/curating contains personally identifiable information or offensive content?
		\answerYes{} \emph{Face information appeared in the FFHQ dataset, the privacy of which is discussed in the cited article.}
	\end{enumerate}
	
	\item If you used crowdsourcing or conducted research with human subjects...
	\begin{enumerate}
		\item Did you include the full text of instructions given to participants and screenshots, if applicable?
		\answerNA{}
		\item Did you describe any potential participant risks, with links to Institutional Review Board (IRB) approvals, if applicable?
		\answerNA{}
		\item Did you include the estimated hourly wage paid to participants and the total amount spent on participant compensation?
		\answerNA{}
	\end{enumerate}
	
\end{enumerate}
\end{document}